\documentclass{article}

    \usepackage[preprint]{neurips_2022}

\usepackage[utf8]{inputenc} 
\usepackage[T1]{fontenc}    
\usepackage{hyperref}       
\usepackage{url}            
\usepackage{booktabs}       
\usepackage{amsfonts}       
\usepackage{nicefrac}       
\usepackage{microtype}      
\usepackage{xcolor}         
\usepackage{graphicx}
\usepackage{float}

\title{DiffuGen: Adaptable Approach for Generating Labeled Image Datasets using Stable Diffusion Models}

\author{
  Michael Shenoda\\
  Department of Computer Science\\
  Drexel University\\
  Philadelphia, PA 19104 \\
  \texttt{michael.shenoda@drexel.edu}\\
  \And
  Edward Kim \\
  Department of Computer Science\\
  Drexel University\\
  Philadelphia, PA 19104 \\
  \texttt{ek826@drexel.edu} \\
}

\begin{document}

\maketitle

\begin{abstract}
    Generating high-quality labeled image datasets is crucial for training accurate and robust machine learning models in the field of computer vision. However, the process of manually labeling real images is often time-consuming and costly. To address these challenges associated with dataset generation, we introduce "DiffuGen," a simple and adaptable approach that harnesses the power of stable diffusion models to create labeled image datasets efficiently. By leveraging stable diffusion models, our approach not only ensures the quality of generated datasets but also provides a versatile solution for label generation. In this paper, we present the methodology behind DiffuGen, which combines the capabilities of diffusion models with two distinct labeling techniques: unsupervised and supervised. Distinctively, DiffuGen employs prompt templating for adaptable image generation and textual inversion to enhance diffusion model capabilities. 
    \begin{figure}[H]
        \centering

            \includegraphics[width=0.16\linewidth]{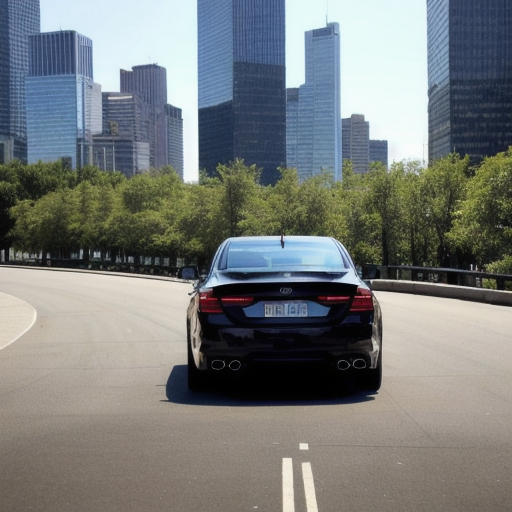}
            \hfill
            \includegraphics[width=0.16\linewidth]{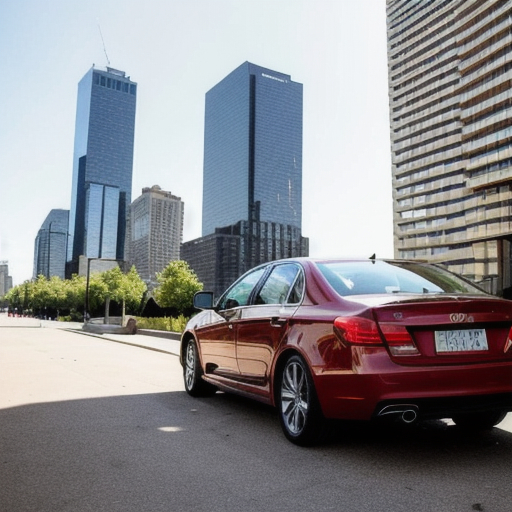}
            \hfill
            \includegraphics[width=0.16\linewidth]{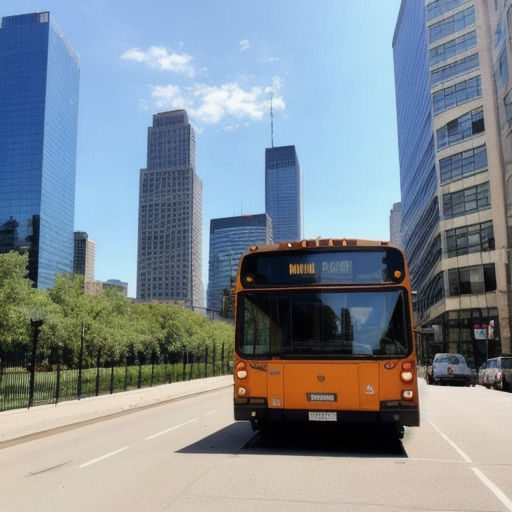}
            \hfill
            \includegraphics[width=0.16\linewidth]{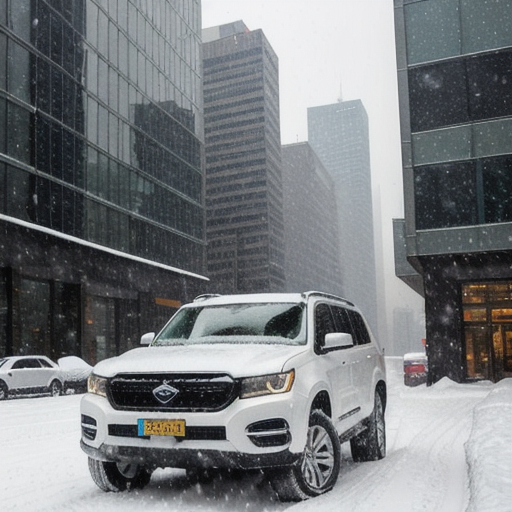}
            \hfill
            \includegraphics[width=0.16\linewidth]{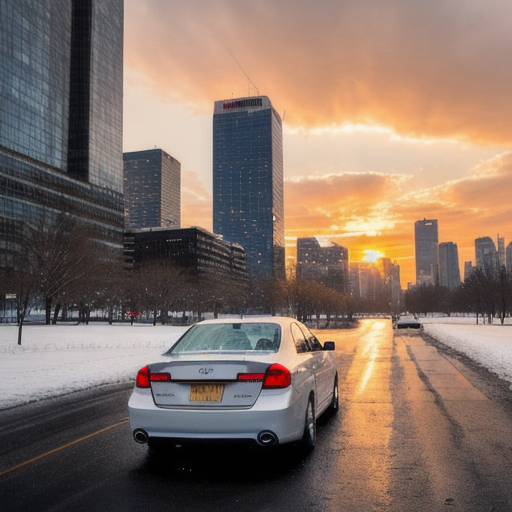}
            \hfill
            \includegraphics[width=0.16\linewidth]{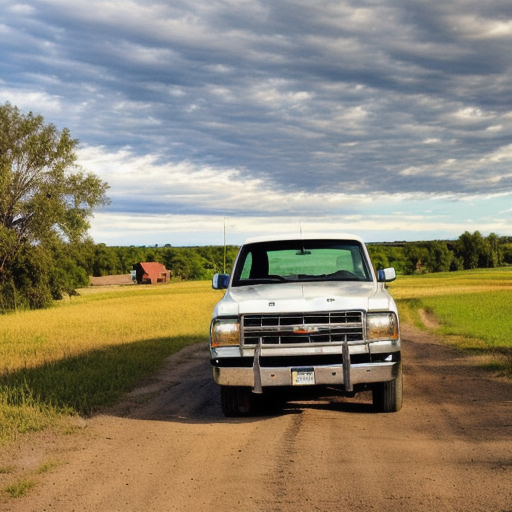}
            \hfill
            \includegraphics[width=0.16\linewidth]{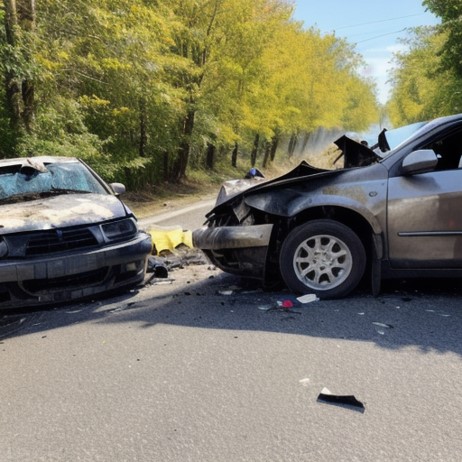}
            \hfill
            \includegraphics[width=0.16\linewidth]{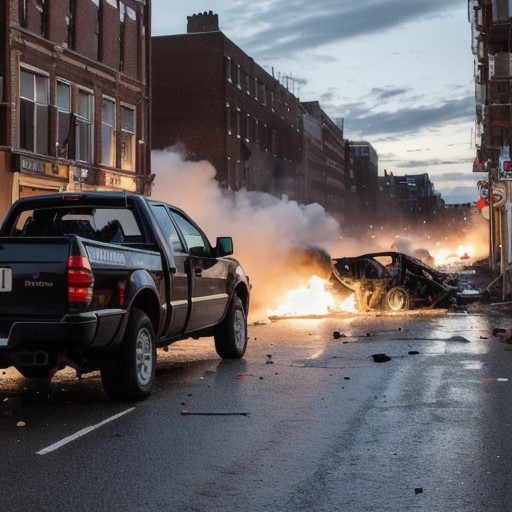}
            \hfill
            \includegraphics[width=0.16\linewidth]{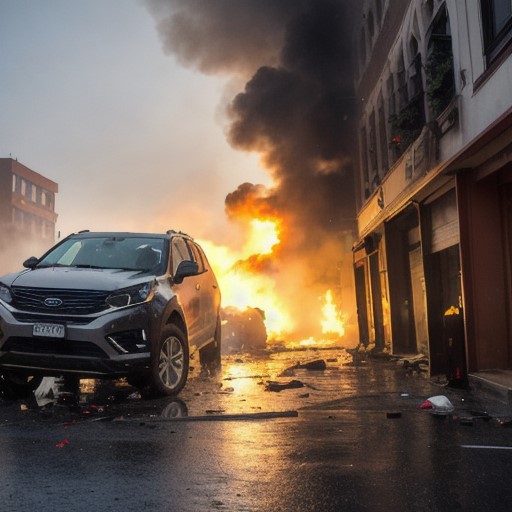}
            \hfill
            \includegraphics[width=0.16\linewidth]{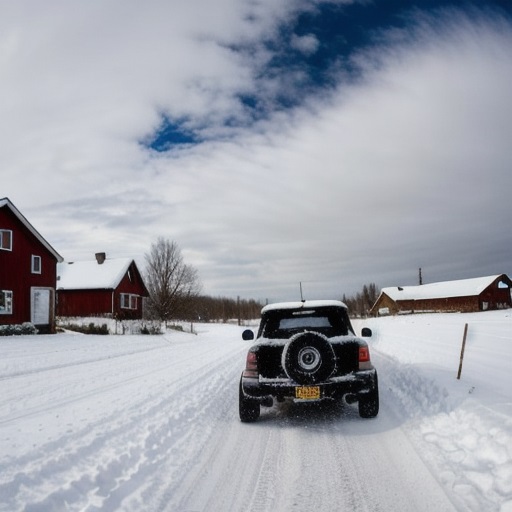}
            \hfill
            \includegraphics[width=0.16\linewidth]{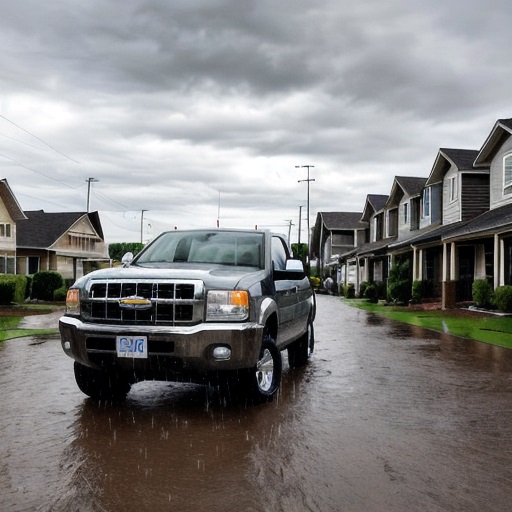}
            \hfill
            \includegraphics[width=0.16\linewidth]{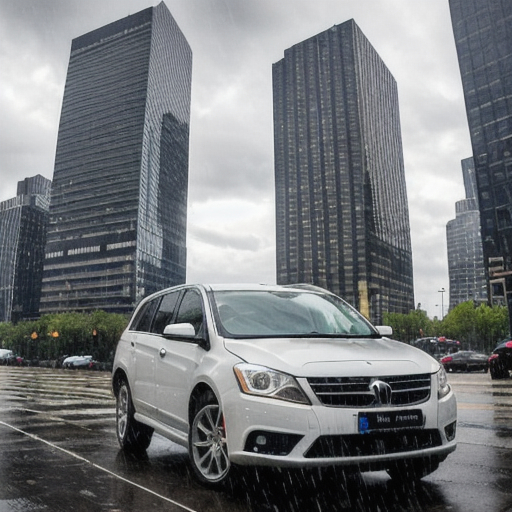}
            \includegraphics[width=1.0\linewidth]{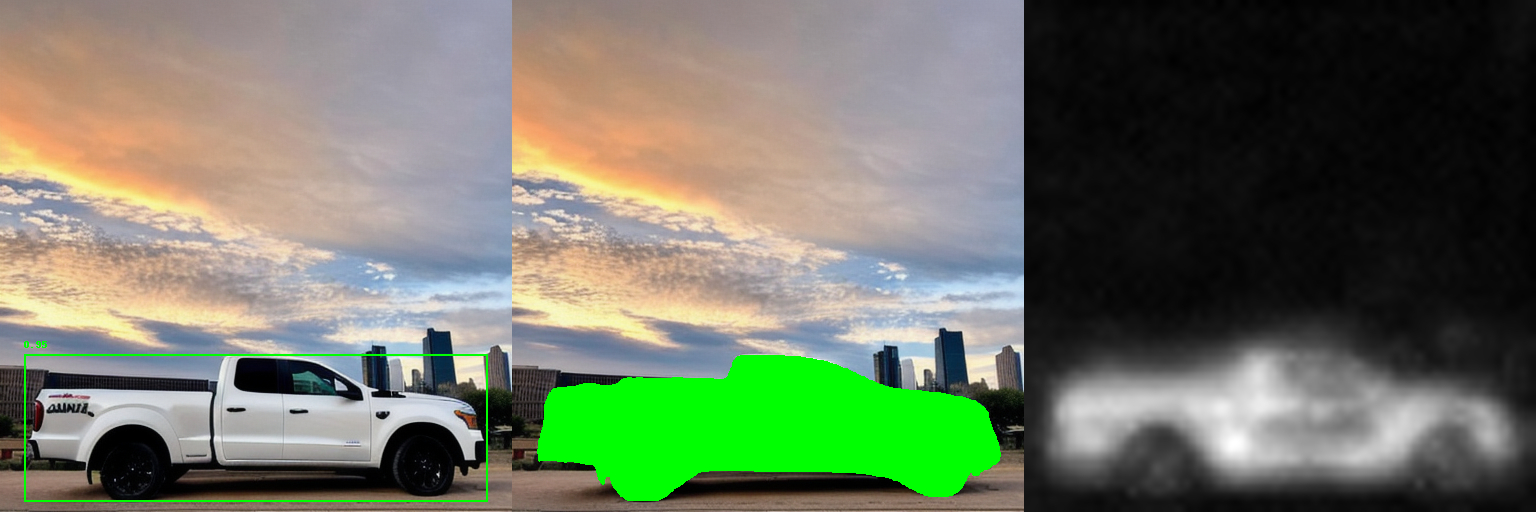}
 
        \caption{Generated diverse images with visualization of the labels and cross attention heatmap}
        \label{fig:abstract_images}
    \end{figure}
\end{abstract}

\section{Introduction}
Generating labeled image datasets for machine learning and computer vision applications is pivotal for model training and evaluation. The quality of these datasets significantly impacts model performance and generalization. In this context, stable diffusion models emerge as a promising avenue for dataset generation due to their ability to generate high-resolution and realistic images. The key objective of this work is to address the challenge of generating diverse and accurately labeled datasets, enabling the development of more robust machine learning models. Through simple techniques such as prompt templating and textual inversion, DiffuGen enhances dataset diversity and generation capabilities. The dual labeling techniques are introduced to compliment each other. The unsupervised method is useful when lacking a supervised model for labeling, where it utilizes the cross attention attribution heatmaps, extracted through the diffusion pipeline, to produce course labels.  The supervised method is effective when an existing image segmentation model exists and ready to be used for labeling and further fine-tuned with generated image datasets. The approach discussed in this paper doesn't require any training, except for expanding a diffusion model with textual inversion. Our experiments showcase the effectiveness of our approach in producing diverse and accurately labeled datasets, offering a promising solution for advancing research in machine learning applications. In our demonstrations we focused on generating cars datasets with ability to generate difficult visual scenarios such as car accidents that involves severe collisions. 

\section{Related Work}

The field of dataset generation has witnessed the advent of various approaches, including GAN-based approaches such as DatasetGAN [3] and BigDatasetGAN [4]. These techniques offer solutions for image synthesis but lack the quality and flexibility of image generation of stable diffusion models. Additionally, the emergence of DiffuMask [9] highlights the potential of stable diffusion models for semantic segmentation but does not cover a broader generation and labeling scope. Our approach distinguishes itself by offering a comprehensive and flexible solution to the labeling challenge by offering semantic segmentation, bounding polygons for instance segmentation and bounding boxes for object detection.

\section{Methodology}
DiffuGen provides a robust framework that integrates pre-trained stable diffusion models, the versatility of prompt templating, and a range of diffusion tasks. By using an input configuration JSON, users can specify parameters to generate image datasets using three primary stable diffusion tasks. Each of these tasks not only benefits from the prompt templating mechanism, ensuring adaptability and richness, but also comes with its dedicated integral labeling pipeline. This design allows DiffuGen to provide both supervised and unsupervised labeling methods tailored to the specific needs of each task, ensuring a well-aligned and efficient labeling process for diverse application needs.

    \begin{figure}[H]
        \centering
        \includegraphics[width=1.0\textwidth]{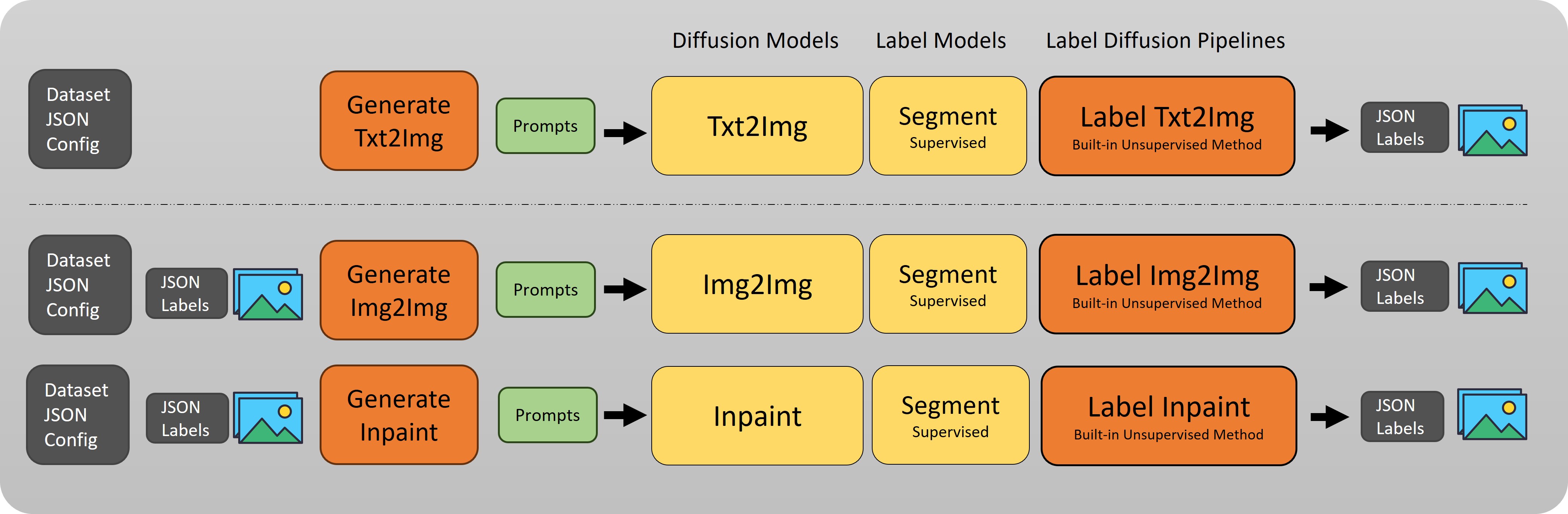}
        \caption{DiffuGen Framework Overview}
        \label{fig:diffugen_overview}
    \end{figure}

\subsection{Utilizing Pre-Trained Stable Diffusion Models}
Pre-trained stable diffusion models serve as the cornerstone of DiffuGen, offering consistency, quality, and adaptability in image generation. During our selection process for the optimal model, we began with the "stable-diffusion-v1-5" [5]. Yet, our evaluations indicated that this model fell short in delivering the desired realism for our dataset generation. The realism is essential to maximize the datasets' relevance in training machine learning models for high-fidelity vision applications. Recognizing these constraints, our exploration led us to the "Realistic\_Vision\_V4.0" [6] model, which distinctly excelled in producing photo-realistic images. Crucially, it retained an accurate object representation with minimal deformation.

\subsection{Prompt Templating}
Prompt templating is the cornerstone of DiffuGen's adaptability and flexibility. Users craft prompt templates populated with replaceable attributes, such as object names, viewpoints, weather conditions, and more. 
This mechanism allows for the reuse of the same prompt to create a multitude of variations, vastly enhancing the diversity of samples generated.While it forms the foundation for the text-to-image task, generating the initial dataset, its utility is not restricted to this phase alone. The same templating mechanism is adeptly reused in the subsequent tasks, image-to-image and inpainting. By allowing the replacement or introduction of attributes, which might not have been specified during the initial text-to-image phase, prompt templating ensures continuous adaptability throughout the dataset generation process.

\subsection{Extending Image Diversity with Different Diffusion Tasks}
Text-to-Image: Serving as the initial phase in dataset creation, this task utilizes prompt templates and their replaceable attributes to generate a diverse set of images. This image dataset set the baseline for the subsequent tasks, providing a rich and diverse foundation dataset.

Image-to-Image: Building upon the foundational dataset from the text-to-image phase, this task introduces variations such as changes in lighting and environment. It benefits from the same prompt templating mechanism, enabling users to modify or introduce new attributes seamlessly, thereby enhancing the diversity and richness of the dataset.

In-painting: Going beyond mere object replacement, this task dive deep into texture variations and color alterations of the object. It is not just about introducing new changes but also for refining them. The prompt templating once again plays a crucial role here, providing users with the ability to specify and guide the changes by introducing newer objects to replace or altering them, resulting in a dataset that's both expansive and detailed.

\subsection{Expanding Stable Diffusion Capability with Textual Inversion}
Textual inversion serves as a powerful technique, enabling the capture of novel concepts from a limited set of example images. This technique holds the potential to significantly enhance the precision and control over image generation in the text-to-image pipeline.

The essence of textual inversion lies in its ability to introduce new "words" into the text encoder's embedding space. These words, learned from a handful of example images, extend the vocabulary of the model where more attention is brought in. This empowers users to drive personalized image generation, steering it towards specific and nuanced visual concepts.

An additional advantage of the textual inversion technique is its lightweight nature, few kb in file size. The learned textual inversion embeddings can be transferred to other models derived from the same base, preserving the capability to control and enrich image generation.

As a demonstration of the textual inversion's technique, we focused on training a rare object that is typically unseen on the road: "grand-piano" as shown in Figure 3. In instances where the original model struggled to generate a piano on the road, the textual inversion successfully refined the generation process. This exemplifies how textual inversion can fill the gaps in object visual representations within diffusion model. In addition, we trained a car-accident textual inversion concept to fine tune examples of car collisions on the road, few samples showing in Figure 1. 

\begin{figure}[H]
    \centering
    \includegraphics[width=1.0\textwidth]{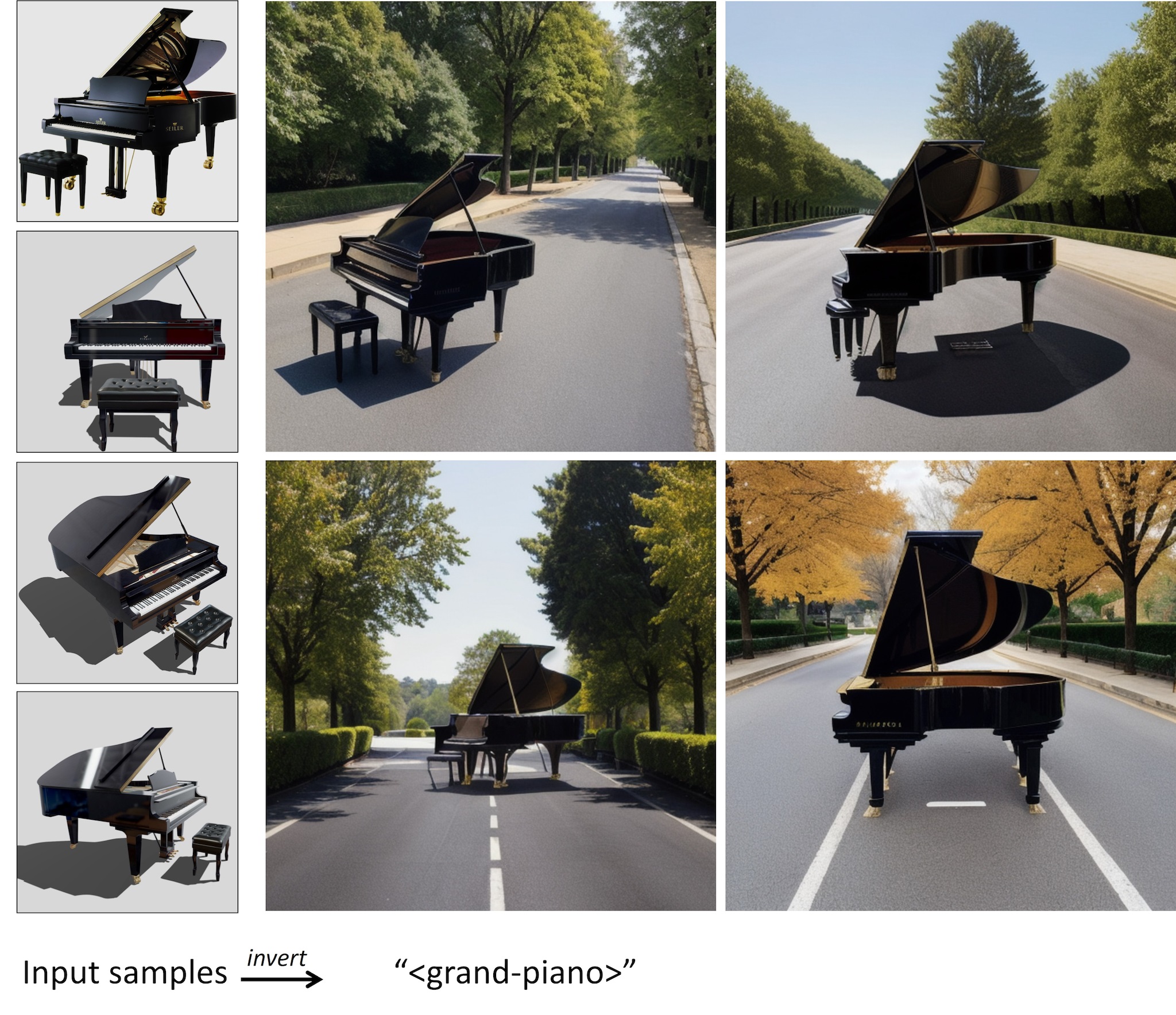}
    \caption{Example of Textual Inversion for <grand-piano> as a rare object concept on road. On the left are some of the real samples used for training, and on the right showing the generated images after training}
    \label{fig:textual_inversion_grand_piano}
\end{figure}

\subsection{Unsupervised Labeling with Cross Attention Attributions}

The approach uses cross attention attributions heatmaps introduced in "What the DAAM: Interpreting Stable Diffusion Using Cross Attention" [1]. These heatmaps offer a visual representation of the relationship between textual prompts and pixel-level influences in the generated images. It works by upscaling and aggregating cross-attention word-pixel scores within the denoising subnetwork of stable diffusion models, resulting in heatmaps that highlight areas influenced by specific words. We build upon this technique to provide a foundation for automatically generating labels, such as semantic mask, bounding polygons, and bounding boxes, without the need for manual annotations.

\subsubsection{Semantic Mask Extraction}
After a cross attention heatmap is obtained, we apply Otsu[2] adpative thresholding on the heatmap image to extract a coarse semantic mask that outline the shapes of objects indicated by the textual prompts. Further refinement is done by performing erosion and dilation oprations on binary mask to segregate close by binary blobs and fill small holes. 

\begin{figure}[H]
    \centering
    \includegraphics[width=0.75\textwidth]{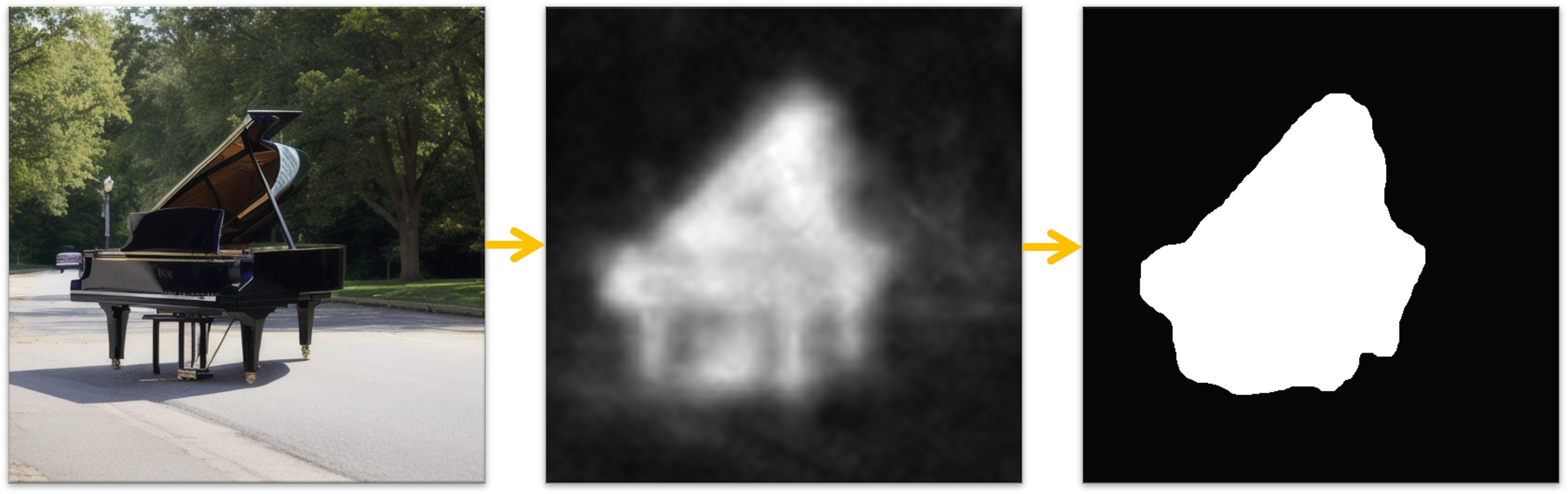}
    \caption{Visualizing semantic mask generation by adaptive thresholding of cross attention heatmap during the diffusion process. Demonstrating the true potential for labeling rare objects or anomalies that yet to be trained in the supervised labeling model}
    \label{fig:unsupervised_semantic_mask_piano}
\end{figure}

\subsubsection{Bounding Polygons and Bounding Box Localization}
We identify contours within the semantic binary mask by using a simple chain approximation. By iterating through the detected contours, we precisely fit bounding boxes around the object-centric regions. This process results in bounding polygon and bounding box coordinates encapsulating the generated objects.
In addition, we introduce a object label scoring mechanism for the generated labels. This scoring method takes into account both the size of the object represented by the bounding box and its intensity level within the cross attention heatmap. This approach provides a unique way to prioritize and evaluate the significance of the generated labels.

\begin{figure}[H]
    \centering
    \includegraphics[width=0.75\textwidth]{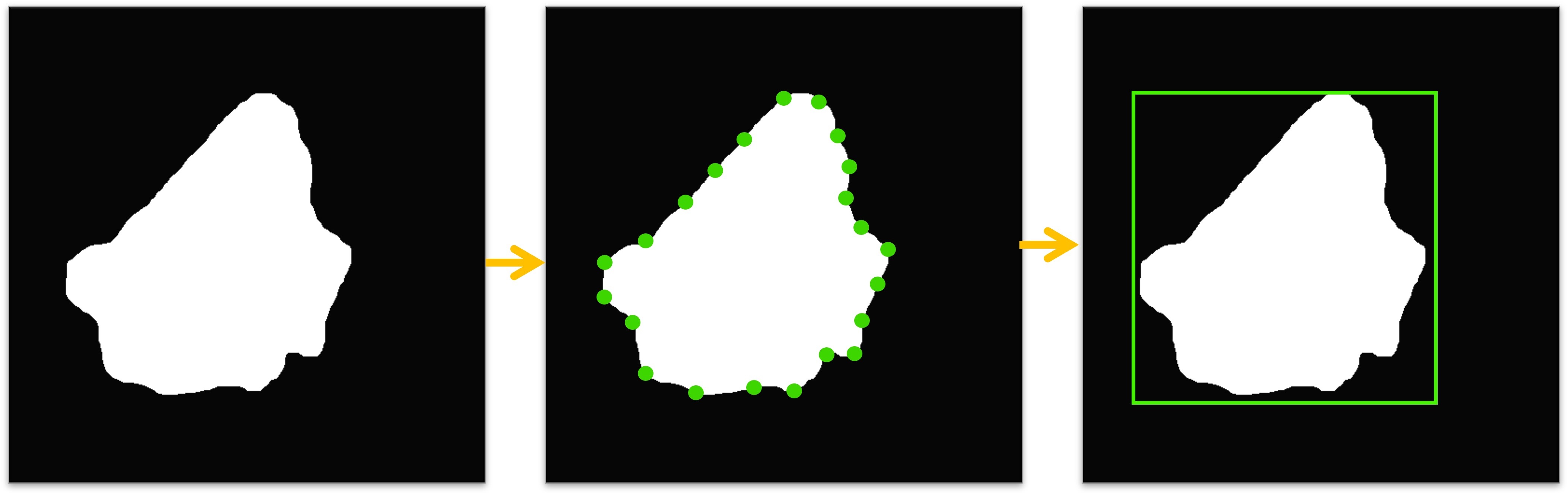}
    \caption{Visualizing bounding polygon and box localization by finding contours of semantic binary mask.}
    \label{fig:unsupervised_bounding_polygon_box_piano}
\end{figure}

\subsection{Supervised Labeling with Existing Segmentation Models}
For scenarios demanding higher precision, DiffuGen proposes the use of supervised segmentation models such as YOLOv8-seg[7] and Mask2Former[8]. In the cases where existing pre-trained models fails to predict the generated images, DiffuGen's unsupervised cross-attention technique can serve as a foundation, facilitating the training of new models using labels produced by the unsupervised approach. Currently, DiffuGen integrates with YOLOv8-seg as a proof of concept to demonstrate the approch. 

\section{Experiments and Results}
To validate the effectiveness of DiffuGen, we undertook a series of experiments. The main objectives were to assess the quality and diversity of the generated images, and the accuracy of the labels provided by the unsupervised and supervised methods.

\subsection{Dataset Generation and Diversity}
Using the text-to-image, image-to-image, and in-painting diffusion task, we generated images of various car scenarios, refer to Figure 1. A majority were normal scenarios while a fraction, specifically controlled by textual inversion, included piano on road and car accidents. Visual assessment showed a high degree of realism, and the diversity in scenarios, colors, lighting conditions, and object placements were commendable.

\begin{figure}[H]
    \centering
    \includegraphics[width=0.25\linewidth]{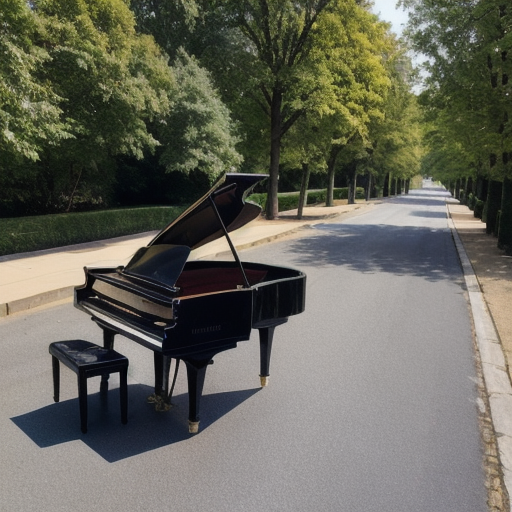}
    \includegraphics[width=0.25\linewidth]{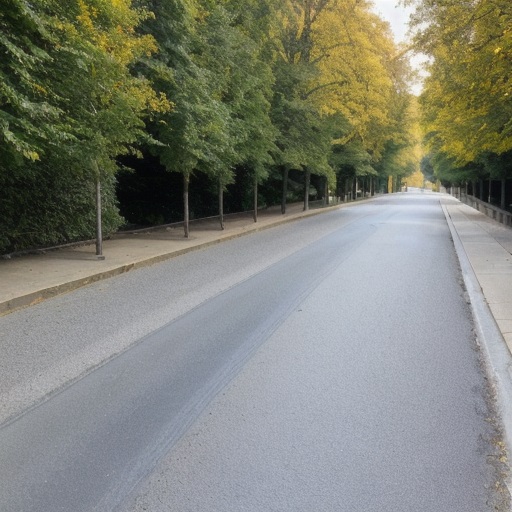}

    \includegraphics[width=0.25\linewidth]{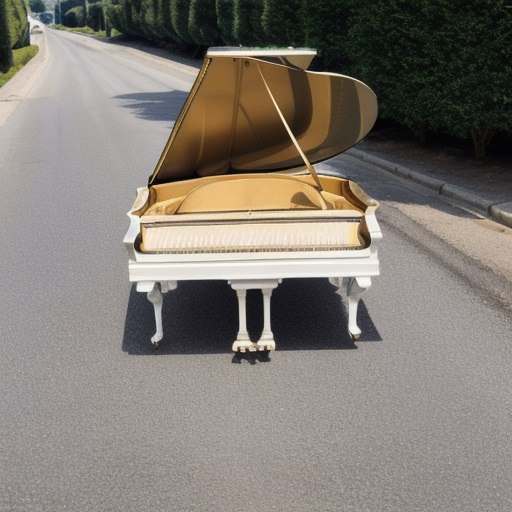}
    \includegraphics[width=0.25\linewidth]{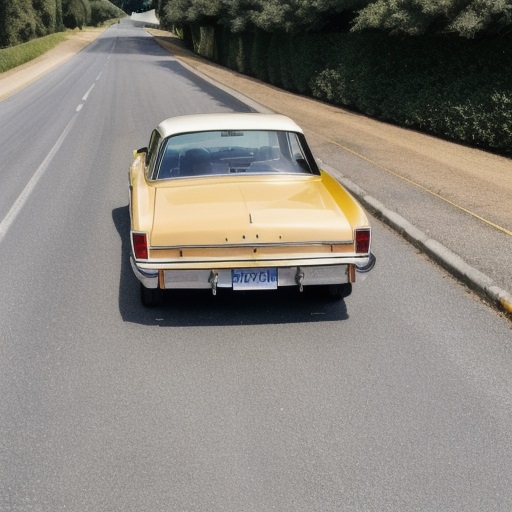}

    \caption{Showing generated images for classic piano on road with textual inversion on left and without on right using same prompt and random seed }
    \label{fig:textual_inversion_side_by_side}
\end{figure}

\subsection{Labeling Accuracy}
Through exterminations, the accuracy of a supervised approach is unmatched, as long as annotated samples exist to begin with. That's where the unsupervised approach would shine to bridge the gap and kick start the supervised approach.

\subsubsection{Unsupervised Labeling}
We visually inspected the labels produced by the unsupervised approach. We found that the majority of the images were labeled accurately, particularly in scenes with fewer objects. In crowded or complex scenes, however, inconsistencies in object detection were occasionally noted. That is due to the inherited limitation of cross attention heatmap approach. 

\subsubsection{Supervised Labeling}
Utilizing YOLOv8-seg, we found the supervised method to have a higher labeling accuracy and finer segmentation boundaries. With exception to the car accidents with severe collision generated with textual inversion. 

\section{Limitations and Future Enhancements}
DiffuGen inherits biases from the underlying diffusion model, impacting generated data. Also, the image quality relies on the model's quality. Relying solely on visual inspection introduces subjectivity. Quantitative assessment is beneficial for future iterations.
Looking ahead, enhancements for DiffuGen include implementing objective metrics to gauge image quality and label accuracy, expanding evaluations across domains, refining unsupervised labeling techniques, and addressing model biases through diverse training data.

\section{Conclusion}
DiffuGen offers a new approach to creating high-quality labeled image datasets. The challenges traditionally associated with manual labeling are greatly reduced, and our visual inspections underline its efficacy. While there is room for improvement, DiffuGen marks a significant stride in the realm of dataset generation, proffering considerable advantages to the computer vision and machine learning domain.

\section*{References}
{
\small
[1] Tang, Raphael, et al. "What the DAAM: Interpreting Stable Diffusion Using Cross Attention." arXiv, 8 Dec. 2022. arXiv.org, \url{https://doi.org/10.48550/arXiv.2210.04885}.

[2] Otsu, Nobuyuki. "A Threshold Selection Method from Gray-Level Histograms," in IEEE Transactions on Systems, Man, and Cybernetics, vol. 9, no. 1, pp. 62-66, Jan. 1979, doi: 10.1109/TSMC.1979.4310076.

[3] Zhang, Yuxuan, et al. DatasetGAN: Efficient Labeled Data Factory with Minimal Human Effort. arXiv, 19 Apr. 2021. arXiv.org, \url{https://doi.org/10.48550/arXiv.2104.06490}.

[4] Li, Daiqing, et al. BigDatasetGAN: Synthesizing ImageNet with Pixel-Wise Annotations. arXiv, 12 Jan. 2022. arXiv.org, \url{https://doi.org/10.48550/arXiv.2201.04684}.

[5] Stable Diffusion V1.5 - Hugging Face, 
\url{https://huggingface.co/runwayml/stable-diffusion-v1-5}.

[6] Realistic Vision V4 - Hugging Face,  \url{https://huggingface.co/SG161222/Realistic_Vision_V4.0}.

[7] Jocher, G., Chaurasia, A., \& Qiu, J. (2023). YOLO by Ultralytics (Version 8.0.0) [Computer software]. \url{https://github.com/ultralytics/ultralytics}.

[8] Cheng, Bowen, et al. Masked-Attention Mask Transformer for Universal Image Segmentation. arXiv, 15 June 2022. arXiv.org, \url{https://doi.org/10.48550/arXiv.2112.01527}.

[9] Wu, Weijia, et al. DiffuMask: Synthesizing Images with Pixel-Level Annotations for Semantic Segmentation Using Diffusion Models. arXiv, 11 Aug. 2023. arXiv.org, \url{https://doi.org/10.48550/arXiv.2303.11681}.
}
\end{document}